# Beyond Defensive Reporting: Machine Learning for Active Anti-Money Laundering Control in Insurance

Dara Goldar
Fremtind Insurance
University of Oslo

Geir Kjetil Ferkingstad Sandve
University of Oslo

Martin Jullum
Norwegian Computing Center

**Abstract**

Money laundering through insurance claims poses a threat to insurers both through fraudulent payouts and reputational and regulatory risk. Despite this, little research has examined how such laundering can be prevented. This paper examines whether machine learning can help insurers flag suspicious claims before payout, shifting the focus from passive reporting to active prevention. Using production data from a major Norwegian insurer, we train gradient-boosted decision tree models to detect claims later reported to authorities for suspected money laundering. Because fraud and laundering may share behavioural patterns, we also examine whether insurance fraud labels can serve as an auxiliary training signal. We compare different learning setups using the Budget-Weighted Capture Rate, a metric introduced in this paper to measure how many laundering cases are captured when only a small share of claims can be manually reviewed. The results show that incorporating fraud-related investigation labels substantially improves laundering detection. The best-performing model captures nearly two-thirds of laundering cases within the top-ranked 2–6 percent of claims selected for investigation. To our knowledge, this is the first empirical study of machine learning for money laundering detection in insurance claims.

## 1. Introduction

Some money laundering techniques incorporate elements of fraud. An example from the insurance sector is an offender insuring a stolen asset and subsequently reporting it damaged. The fraud occurs when the offender lies about the asset's origin to obtain a payout. That payout then serves as a laundering step, replacing illegal property with seemingly legitimate funds.

Such money laundering schemes pose a dual threat to insurers. Paying out fraudulent claims reduces profits, and facilitating money laundering can lead to reputational damage and regulatory sanctions (Gottschalk, 2024; Insurance Authority, 2024). To mitigate these risks, insurers rely on automated systems to assess risk, screen claims, and report suspicious activity. Such systems require continuous tuning to remain effective (FATF, 2014; KPMG, 2014).

Achieving effective calibration of monitoring systems is challenging. Evidence from AML in banking suggests that experienced professionals exhibit cognitive biases in risk assessment, performing no better than novices when compared to court outcomes (Ogbeide et al., 2023). Such biases can influence how thresholds are set in practice. Moreover, rules-based systems with fixed thresholds struggle to distinguish legitimate from suspicious behavior, especially as patterns change over time. These limitations have motivated the use of machine learning (Oztas et al., 2024).

Recent work applies both supervised and unsupervised methods to money laundering detection (Mousavian & Miah, 2025). Supervised approaches generally perform better when labelled data are available (Ahmad & Quegan, 2013; Khoei & Kaabouch, 2023; Sur et al., 2023). A key challenge in money laundering detection is the extreme class imbalance and limited availability of valid labels (Kennedy et al., 2023). Ideally, models would be trained on a substantial number of court-confirmed money laundering cases, but such examples are rare in practice. As a result, models are typically trained to predict suspicious activity reports (SARs) rather than confirmed crime (Jullum et al., 2020; Hayble-Gomes, 2023; Johannesen & Jullum, 2025).

Optimising SAR filings with machine learning may improve compliance efficiency, but does not necessarily reduce financial crime. A review in the UK found that, in the absence of clear feedback, financial institutions tend to engage in so-called defensive reporting, submitting reports mainly to show compliance (Law Commission, 2019), while in the US, the overall impact of the reporting regime remains difficult to assess (U.S. Government Accountability Office, 2024). Despite limited evidence of effectiveness, the AML regime persists partly because it allows governments to signal action against hidden threats, and because it supports geopolitical objectives such as sanctions enforcement against countries like North Korea, Russia and Iran (Levi, 2019).

Given this lack of measurable impact, a pragmatic alternative is to focus on a subset of SARs where suspicion leads to direct intervention, typically by stopping the flow of suspicious funds. In doing so, the reporting entity shifts from a passive reporter to an active preventer. Though not proven in court as money laundering, targeting cases where direct intervention is possible provides measurable impact rather than just compliance efficiency.

This study focuses on such intervention in insurance claims. Because of tipping-off rules, insurers cannot deny a claim based solely on money laundering suspicion, but they can refuse the payout if they also suspect insurance fraud (FATF, 2025). To systematically identify these overlapping cases, we train machine learning models on a large dataset of claims provided by Fremtind Insurance, one of Norway's largest insurers. Appendix A details three investigations from our dataset, including organised vehicle identity theft, the concealment of illicit income, and the illegal diversion of corporate assets, to illustrate what this overlap of fraud and laundering suspicion looks like in practice.

We also examine whether including additional insurance fraud claims during training improves the detection of laundering. The laundering cases in our data already involve suspected fraud and denied payouts. However, fraud claims without recorded laundering suspicion are roughly ten times more frequent. Because insurance fraud shares important typologies with money laundering (U.S. Department of the Treasury, 2024), this larger pool of fraud cases may help compensate for the small number of laundering cases. To assess this, we compare alternative learning formulations, including binary and multiclass models.

Our contribution to the literature on financial crime prevention is threefold. First, we present, to the best of our knowledge, the first empirical study of machine learning for detecting money laundering in the insurance sector. Second, we shift model optimisation from report generation toward intervention through claim denial. Third, we demonstrate that incorporating related fraud labels through alternative learning formulations improves laundering detection under extreme class imbalance.

The remainder of this paper is structured as follows: Section 2 begins by clarifying the target label and what the considered company (Fremtind) regards as suspected money laundering in insurance claims. We then describe the dataset and present the machine learning framework used for detection and evaluation. Section 3 reports empirical results, and Section 4 discusses implications for prevention-oriented AML design and the limitations associated with institutionally constructed laundering labels.

## 2. Methods

### *2.1 Claims investigation process*

All observations in this study originate from registered insurance claims. The process starts with a claim submission, where 99.9% of claims are resolved without investigation and treated as legitimate.

A small fraction of claims is selected for insurance fraud investigation. Insurance fraud refers to intentionally obtaining or attempting to obtain a claim payout through false or misleading information (Norwegian Penal Code § 375, 2005). These investigations yield three possible outcomes: the suspicion is dismissed (claim treated as legitimate); suspicion remains but evidence is insufficient for police reporting (classified as a fraud investigation); or evidence is sufficient to report to the police (classified as fraud).

A subset of fraud investigations trigger a money laundering investigation. Money laundering refers to handling proceeds derived from criminal activity (Norwegian Penal Code § 337, 2005). Investigations result in either dismissed suspicion or a Suspicious Activity Report (SAR), where the reporting threshold requires that suspicion remains after the internal money laundering investigation (Norwegian Ministry of Finance, 2018). In Fremtind's investigative practice, typical SARs originating from insurance claims involve assets linked to theft, high-value imports lacking customs declaration, or assets with unknown origin and of disproportionate value relative to the policyholder's income.

Figure 1 illustrates the investigation process, while Table 1 defines the resulting class mapping to target classes used for machine learning prediction. Appendix A provides three anonymised investigations to demonstrate how such cases arise in practice. The investigations are drawn from the set of claims the machine learning models are designed to detect and are intended to provide some transparency regarding the evidentiary basis underlying the money laundering label used in this study.

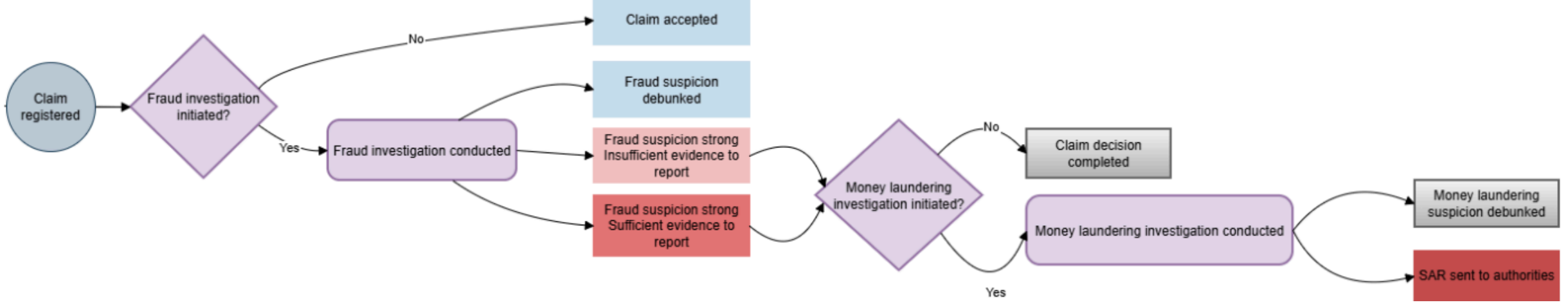

***Figure 1:*** *The process starts with a registered insurance claim (from left). Diamonds denote decision nodes to initiate an investigation, rounded rectangles in purple denote investigative processes, and regular rectangles denote investigation outcomes. A fraud investigation may trigger a money laundering investigation, which can result in a suspicious activity report (SAR) to authorities. Class labels are assigned based on the final investigation outcomes.*

| **Fraud investigation outcome** | **Laundering investigation outcome** | **Target class** |
|---|---|---|
| No investigation (Claim accepted) | No investigation | Legitimate |
| Fraud suspicion debunked | No investigation | Legitimate |
| Fraud suspicion strong – insufficient evidence to report | No investigation | Fraud investigation |
| Fraud suspicion strong – sufficient evidence to report | No investigation | Fraud |
| Fraud suspicion strong – insufficient evidence to report | Money laundering suspicion debunked | Fraud investigation |
| Fraud suspicion strong – sufficient evidence to report | Money laundering suspicion debunked | Fraud |
| Fraud suspicion strong – insufficient evidence to report | SAR sent to authorities | **Laundering** |
| Fraud suspicion strong – sufficient evidence to report | SAR sent to authorities | **Laundering** |

***Table 1:*** *Fraud investigation outcome and laundering investigation outcome together define what target class a claim falls in. SAR sent to authorities always takes precedence. This ensures a multi-class set-up.*

### *2.2 Dataset construction*

Insurance claims data were obtained from the enterprise data warehouse of Fremtind Insurance. The dataset covers insurance claims related to individuals from January 2020 through December 2025. It is a subset of claims in that period and was chosen by Fremtind for research purposes. The dataset consists of four mutually exclusive classes of claims:

- Reported money laundering claims: 148 claims in which Fremtind denied a claim and reported the customer to authorities. These claims were also investigated for insurance fraud. The fraud investigation either resulted in a formal report to the police or was closed due to lack of evidence. In all cases, the insurance claim was rejected. Hereafter, this class is termed *laundering*.
- Insurance fraud claims: 1,582 claims concluded as insurance fraud following internal investigation. None of these claims were investigated for laundering. Consequently, there is no overlap with the laundering class. Hereafter, this class is termed *fraud*.
- Insurance fraud investigation claims: 1,800 claims investigated for fraud but closed due to insufficient evidence. Although not escalated to police, specialists identified fraud indicators in these cases, making them typologically similar to the fraud class and useful for model training. Notably, the internal code used to flag these cases was applied more frequently over time. This category does not overlap with any other. Hereafter, this class is termed *fraud investigation.*
- Legitimate claims: Approximately 2,500,000 claims belonging to 850,000 customers that have never been investigated for insurance fraud or money laundering. Hereafter, this class is termed *legitimate.*

Given the study's focus on direct intervention at the claims stage, the laundering class is restricted to cases where the insurer denied a claim and filed a SAR based on that specific claim. We entirely exclude customers reported for suspected money laundering without any associated suspicious claim to avoid introducing label noise in two directions. First, including paid-out claims from these customers in the laundering class could weaken class separability, as the claims themselves lack the suspicious patterns the model should learn. Second, keeping the same claims in the legitimate class could contaminate the baseline with investigation artefacts and background SAR indicators that do not reflect a standard, unaffected claims process.

The feature set consists of features constructed from information available immediately after claim registration, including claim characteristics and system-generated reservation transactions derived from how the claim is submitted. No features rely on outcomes of fraud or laundering investigations, or on any information generated after claim registration. All features are drawn from the same underlying data sources that support the insurer's existing rule-based monitoring system. In total, the feature set comprises 66 features, of which 9 are categorical and the remainder are numerical. The features can be summarised into the following eight groups:

- Number and type of products, capturing the customer's insurance products at claim registration and historically (16 variables)
- Historical claims information, summarising previous claims, claim-related transactions, and payout behavior prior to the focal claim (7 variables)
- Focal claim information, including claim characteristics, timing, and system-generated reservation transactions derived from how the claim is submitted (11 variables)
- Premium payment history, summarising premium amounts by product type and over historical time windows before the focal claim (12 variables)
- Customer contact information, reflecting how the customer interacts with the insurer (5 variables)
- Financial stress indicators, capturing signals related to payment remarks and financial obligations (1 variable)
- Demographic information, including age, gender, income, wealth, and household composition (13 variables)
- Customer roles in organisations, describing formal roles and affiliations with organisations (1 variable)

Significant care is taken in dataset construction to avoid giving the model information that would not be available in a production environment when assessing new claims, and to reduce the risk that it learns patterns that do not generalise well to future claims. To reduce dependence between observations, each customer is associated with a single claim. This prevents information from the same customer appearing across training and evaluation samples and avoids overly optimistic performance estimates (Rosenblatt et al., 2024). For customers in the investigated classes, most are linked to a single investigated claim. In the rare cases where a customer has multiple investigated claims, we retain the first. This avoids artefacts from repeated investigative handling. For legitimate customers, many have multiple claims during the observation period. One claim per customer is selected uniformly at random.

A further challenge stems from the distribution of investigated claims varying over time relative to the full claim portfolio. These temporal fluctuations in the positive classes may be driven in part by external factors, such as changes in investigative focus, regulatory pressure, or staffing, rather than intrinsic characteristics of individual claims. This can lead to what is known as shortcut-learning: if certain patterns in the features are associated with particular periods, the model may learn to use those patterns as proxies for class membership rather than what differentiates suspicious from legitimate claims. Shortcut learning can make models less reliable when they are applied to new data that differ from the training data (Geirhos et al., 2020). To reduce this risk, we subsample legitimate claims in periods with few investigated cases to better align the temporal distribution of the classes. Figure 2 illustrates the resulting distributions of such subsampling.

To enable robustness assessment, we generate four versions of the dataset using different random seeds that vary the random selections in our dataset construction. Across these versions, laundering, fraud, and fraud investigation claims remain fixed, while the legitimate class varies through the selection of one claim per legitimate customer, as well as through the selection of legitimate claims retained when subsampling to satisfy the temporal balancing constraint.

*2.3 Three machine learning setups for detecting suspicious claims*

We consider two binary and one multiclass learning formulation that differ in how investigation outcomes are represented in the training targets. In the binary setups, outcomes are collapsed into a single positive class contrasted against a negative class, and the model estimates the probability that a claim belongs to the positive class. In the multiclass setup, each investigation outcome is modeled as a distinct class, and the model outputs a probability distribution over all classes. Formally, for each claim $i$ with features $x_i \in R^d$ and target $Y_i$, we have $Y_i \in \{0, 1\}$ in the binary case and $Y_i \in \{0, 1, 2, 3\}$ in the multiclass case. Models are trained using logistic loss in the binary setting and the corresponding multiclass logistic loss in the multiclass setting.

Model selection is based on average precision. For the Binary–Laundering and Multiclass setup, models are selected to maximise average precision on laundering cases. For the Binary–Fraud setup, model selection is based on average precision over all investigated claims. Table 2 summarises the three learning setups.

| **Setup** | **Learning type** | **Target definition $Y_i$** | **Positive class** | **Model selection** |
|---|---|---|---|---|
| Binary-Laundering | Binary | Yi∈ {0,1} | Laundering (SAR sent) | Average precision on $Y_i = 1$ |
| Binary-Investigation | Binary | Yi∈ {0,1} | Fraud investigation, fraud, or laundering | Average precision on $Y_i = 1$ |
| Multiclass | Multiclass | Yi∈ {0,1,2,3} | Legitimate (0), fraud investigation (1), fraud (2), laundering (3) | Average precision on $Y_i = 3$ |

***Table 2:*** *Comparison of the Binary–Laundering, Binary–Investigation, and Multiclass learning setups, detailing how the target variable is defined and how models from inner folds are selected in each experiment.*

In all setups, we employ gradient-boosted decision trees implemented using XGBoost. Tree-based ensemble methods remain state-of-the-art for tabular prediction tasks and have been widely applied in AML research (Grinsztajn et al., 2022; Jullum et al., 2020; Reite et al., 2025). Gradient-boosted trees build an ensemble of shallow trees sequentially so that each learner targets the residuals of the previous ones (Chen & Guestrin, 2016).

We also report XGBoost gain-based feature importance to give an indication of which groups of features the model uses when ranking claims by suspicion. In simplified terms, a feature receives higher importance when that feature helps the model better separate investigated from legitimate claims. Technically, gain measures the improvement in the model's training objective when a feature is used in a tree split (Chen & Guestrin, 2016). Features with higher gain therefore have greater influence on whether a claim receives a higher or lower risk score. We group individual features into the feature groups described in Section 2.2 and sum the gain values within each group. The reported group-level values are then normalised to sum to 100%.

To simulate real-world deployment, models are evaluated using a nested time-series cross-validation procedure where laundering claims define the time splits, implemented using the time-series split functionality in scikit-learn (Pedregosa et al., 2011). Laundering claims are ordered chronologically and partitioned into folds, with cutoff dates for training, validation, and test sets defined by the first and last laundering claim dates in each fold. All other claims are assigned based on their claim dates, such that they fall within the corresponding laundering-defined cutoff intervals, ensuring consecutive, non-overlapping splits that use only information available at the time. Figure 2 illustrates such a split.

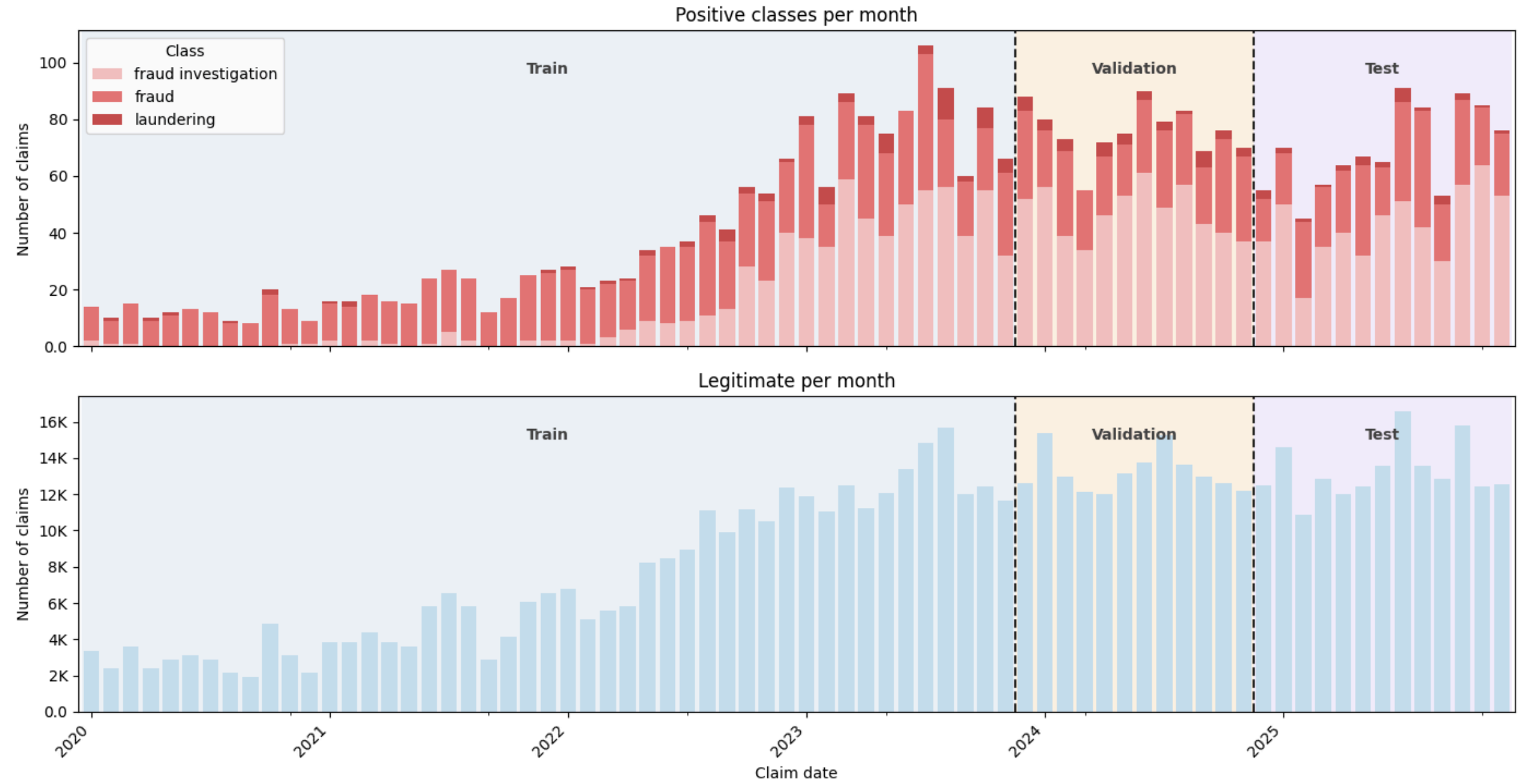

***Figure 2:*** *Schematic of the time-series nested CV pipeline. The dataset shown corresponds to the two learning setups using all investigated claims, using dataset realisation 1, outer fold 3, and inner fold 1. Each claim in the figure is associated with a unique customer and the distribution of dates are roughly matched between the positive classes and the legitimate class.*

Hyperparameters are selected within the inner folds of the nested cross-validation procedure. We tune standard XGBoost parameters, including learning rate, tree depth, and ensemble size, and apply early stopping based on validation log loss. Class imbalance is handled using observation-level sample weights, treated as tunable parameters. In the binary setups, a single weight controls the importance of the positive class, while in the multiclass setup, separate weights are assigned to laundering claims and to the fraud-related classes. Each learning setup evaluates the same number of configurations, combining model and weighting parameters. Appendix B provides the hyperparameter grid.

Using the selection objective defined in Table 2, the best-performing models from the inner loop are combined into an outer-fold ensemble by averaging their predicted probabilities. Let $p_i^{(m)}$ denote the predicted probability produced by inner model $m \in \{1, 2\}$ for claim $i$. The outer-fold ensemble prediction is then given by

$$p_i = \frac{1}{2} \sum_{m=1}^{2} p_i^{(m)}$$

The outer-fold ensemble constitutes a second level of ensembling on top of the gradient-boosted tree ensembles used within each individual model.

The nested cross-validation pipeline is repeated across the four dataset versions constructed as described in Section 2.2. These versions differ only in the pseudorandom selection of one claim per legitimate customer, while all laundering, fraud, and fraud investigation claims remain fixed.

*2.4 Evaluation*

Model performance is evaluated using the Budget-Weighted Capture Rate (BWCR), a metric we introduce to reflect detection effectiveness under constrained investigation capacity. Let *Recall@b* denote the proportion of all true laundering cases appearing among the top $b$ of the ranked portfolio. The Budget-Weighted Capture Rate is defined as

$$BWCR = 0.25 \cdot Recall@2\% + 0.50 \cdot Recall@4\% + 0.25 \cdot Recall@6\%$$

Using multiple cutoffs reduces sensitivity to any single investigation threshold and improves stability. The insurer does not want to disclose its exact capacity constraints but considers the range 2–6% to be realistic. These cutoffs can be interpreted as conservative, moderate, and expansive investigation regimes, with the most emphasis placed on 4% as a mid-range scenario. This focus on the top of the ranked portfolio is consistent with information retrieval practice, where evaluation emphasises the highest-ranked results that a user can realistically examine, for example through metrics such as *Precision@k* (Manning et al., 2008). In our setting, the investigator plays an analogous role, reviewing only a limited portion of the ranked claims.

While BWCR focuses on ranking quality at the top of the portfolio and serves as our primary metric, we also report Average Precision (AP) for the laundering class as a complementary measure of global ranking performance. AP summarises the precision–recall curve and captures how well laundering cases are ranked above non-laundering cases across all thresholds. When BWCR is similar across models, a higher AP indicates a more consistent overall ranking.

## 3. Results

In this section, we examine machine learning methods for detecting claims who are later reported to authorities for money laundering. We also investigate what features are used the most to separate legitimate and suspicious claims.

Table 3 reports the aggregate performance of the three learning setups, presenting the mean and standard deviation of BWCR and AP across 16 evaluations per learning setup. The results show that all three models succeed in surfacing claims related to money laundering within the investigation budget. Despite the very small number of laundering observations relative to legitimate claims, even the Binary-Laundering model achieves a mean BWCR of 53.2 percent, meaning that above half of the laundering cases are captured within the top-ranked portion of claims selected for review. By contrast, a random ordering would be expected to capture only the reviewed share itself, so that BWCR across the 2–6 percent would be expected to capture 4 percent of the laundering cases.

Because earlier outer folds contain fewer laundering cases and therefore less training data, the reported standard deviations reflect not only random variation across runs but also systematic differences between folds arising from unequal training-set sizes.

| ***Learning setup*** | ***BWCR (%)*** | ***Average precision (%)*** |
|---|---|---|
| *Binary-Laundering* | *53.18±8.34* | *1.42±1.20* |
| *Binary-Investigation* | *64.76±6.61* | *2.34±1.28* |
| *Multiclass* | *57.81±8.17* | *2.18±1.56* |

***Table 3****: Table 3 reports the mean and standard deviation of BWCR and Average precision, both evaluated on the laundering class, across 16 evaluations per experiment. Binary-Investigation achieves the highest BWCR and Average precision.*

Moving from the Binary-Laundering setup to the Binary-Investigation setup, where all investigated claims are treated as positive, leads to a clear improvement in BWCR. The average gain is approximately 12 percentage points.

| *Learning setup* | *Fold 0* | *Fold 1* | *Fold 2* | *Fold 3* |
|---|---|---|---|---|
| *Binary–Laundering* | *46.32±4.90* | *48.28±0.70* | *52.37±2.93* | *65.73±3.02* |
| *Binary–Investigation* | *57.33±1.49* | *73.92±2.27* | *60.99±0.83* | *66.81±0.50* |
| *Multiclass* | *46.98±1.49* | *57.11±2.93* | *59.05±2.28* | *68.10±4.16* |

**Table 4**: *Table 4 reports BWCR per fold. Values are means over four dataset realisations with standard deviations. Binary-Investigation performs best in three of four folds, while Binary-Laundering performs worst in all.*

The Multiclass setup improves on Binary-Laundering as measured by BWCR, but does not match the performance of Binary-Investigation. Multiclass outperforms Binary-Laundering in all four outer folds and achieves the highest BWCR in the final fold, while Binary-Investigation performs best in the three earlier folds.

For average precision, the Multiclass setup also improves on Binary-Laundering, but remains slightly below Binary-Investigation. Its mean average precision is 2.18 percent, compared with 1.42 percent for Binary-Laundering and 2.34 percent for Binary-Investigation.

In practical terms, these results demonstrate the viability of automatic triaging immediately upon claim registration. The Binary-Investigation model captures almost two-thirds of laundering cases when only the top-ranked 2–6 percent of claims are selected for review. It also outperforms the insurer's existing rules-based engine for registration-time risk assessment, although the insurer does not disclose the margin of improvement.

Crucially, this means that investigative effort can be concentrated on a small, high-risk subset of claims while intervention is still possible. Because the ranking is produced immediately after registration, flagged claims can be reviewed before payout. Where there is a basis for denying the claim, payment can be stopped; where suspicion of money laundering remains after the internal investigation, a SAR can be filed.

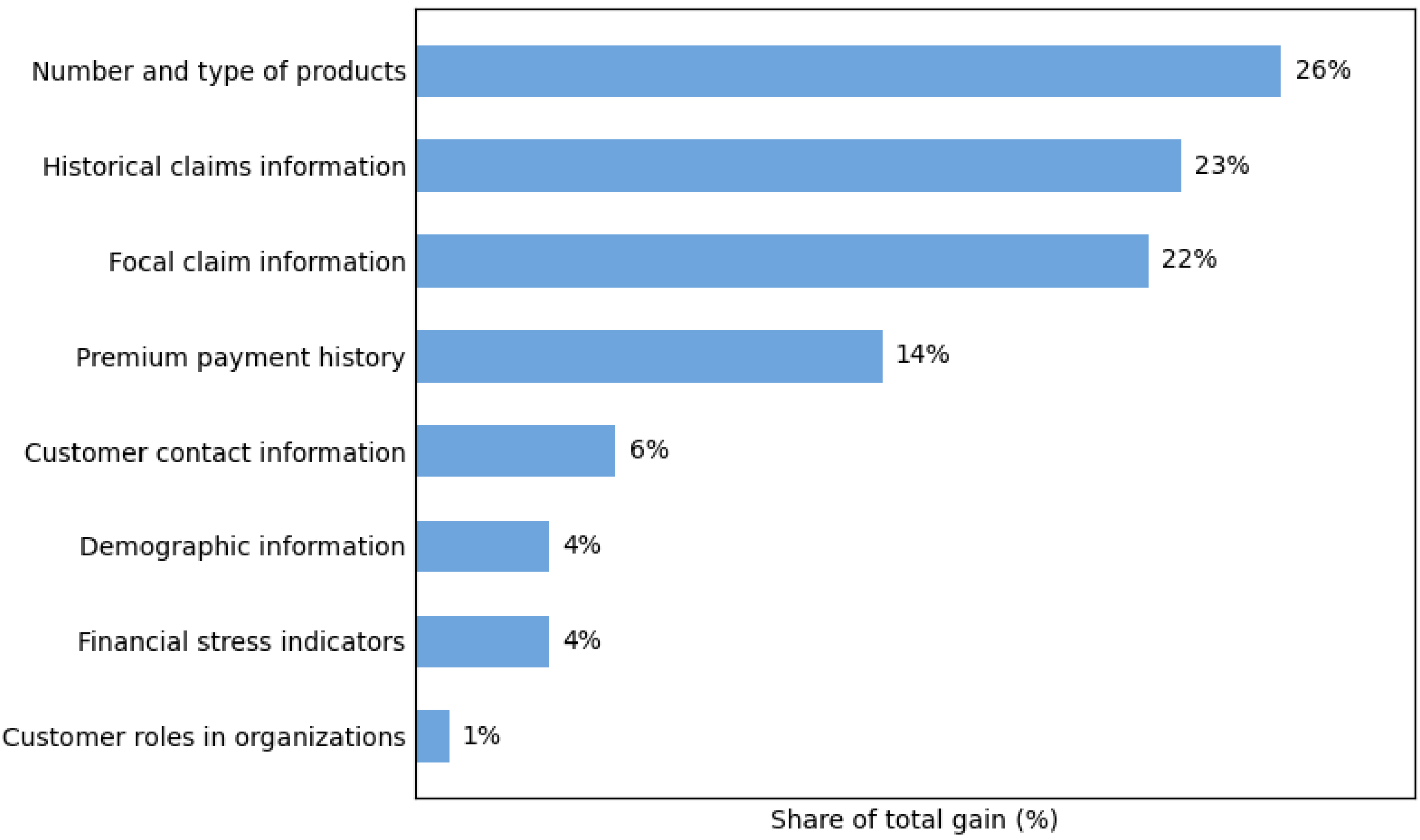

***Figure 3:*** *XGBoost gain-based feature importance by feature group for Binary-Investigation. Values show each group's share of total gain and sum to 100%. Feature groups and the number of features in each group are described in Section 2.2.*

Figure 3 shows which feature groups contribute most to separating legitimate claims from investigated claims in the Binary-Investigation setup. The model relies mainly on information related to insurance activity, particularly product holdings, claims history, the focal claim, and premium payments, to separate investigated claims from legitimate. These together account for 85% of the gain. Demographic and broader customer information contribute substantially less.

## 4. Discussion

This study compares three alternative formulations of the laundering detection problem and shows that target definition affects operational performance as measured by the Budget-Weighted Capture Rate (BWCR), our primary metric reflecting constrained investigative capacity. Training exclusively on laundering labels yields the weakest results. Expanding the positive class to include all investigated claims substantially improves BWCR. Changing to a multiclass model does not improve performance as compared to the simpler Binary-Investigation formulation.

Binary-Investigation also shows the strongest performance as measured by average precision. However, because this metric reflects ordering across the entire portfolio rather than within realistic investigation limits, its practical relevance is secondary to BWCR.

The improvement in BWCR when incorporating related fraud labels likely reflects shared structure between laundering, fraud, and fraud investigation claims. When trained solely on laundering, the model learns from very few positive examples, limiting generalisation. Including all investigated claims provides a larger pool of suspicious cases, improving the early ranking of laundering claims. Allowing the model to distinguish between different positive outcomes in a multiclass setting improves performance relative to Binary-Laundering, but is worse than simply merging all investigated claims into a single positive class. This suggests that the main gain comes from exposing the model to a broader set of suspicious claims.

Beyond the three setups explored here, one could imagine a two-stage detection approach in which a first model identifies claims requiring investigation and a second model detects laundering among those sent for investigation. In practice, however, fraud and laundering investigations are handled by the same investigation team. Once a claim is escalated for investigation, it is examined thoroughly, making it very likely that any potential laundering would be identified as part of that process. The first stage of such a setup would effectively correspond to the Binary-Investigation formulation, making a separate second stage redundant.

The feature importance results suggest that the models rely mainly on insurance activity rather than broader customer background information. Product holdings, premium payments, claims history, and characteristics of the focal claim dominate, accounting for 85% of the total gain. However, this distinction should not be interpreted as a strict separation between insurance activity and customer background. Insurance activity may already correlate with, and indirectly capture, aspects of customers' demographics and financial situation.

Operationally, the models can function as an early triage system for manual claim handling. At present, decisions about which claims to escalate rely heavily on a small number of experienced handlers who use professional judgment to assess suspicion. This makes the process sensitive to availability and staff turnover. A model that ranks all incoming claims allows investigative resources to be concentrated on the highest-risk cases across the full portfolio and provides greater consistency over time. Within a fixed investigation budget, improved ranking enables more high-risk payouts to be identified and potentially stopped before payouts are made.

The results should be interpreted with caution. Laundering cases in this study are defined by internal decisions to deny a claim and submit a suspicious activity report. Based on publicly available court records, none of the reported cases in this dataset have resulted in prosecution for money laundering (Appendix D). A closer examination of three example investigations in Appendix A shows that the laundering labels vary in evidentiary strength and seriousness. Investigation 1 involved a direct technical match to a vehicle reported as stolen, providing a strong indication of organised vehicle crime. Investigation 2 relied primarily on financial inconsistencies and was concluded without further investigative steps that could have strengthened the supporting evidence. Investigations 1 and 2 concerned substantial amounts that would likely qualify as aggravated money laundering under prosecutorial guidelines, whereas investigation 3 involved a comparatively small amount likely to result in a minor financial sanction. In sum, the laundering labels are based on internal conclusions reached after limited investigation, conducted without informing the customer and without systematic feedback from authorities.

These contributions extend earlier work in related domains. Academic literature on money laundering through insurance claims remains scarce and is largely limited to high-level typologies based on FATF case material that is over twenty years old, or to hypothetical laundering techniques (Thanasegaran & Shanmugam, 2008; Bagaskara & Yanti, 2025). Appendix A adds recent case material and documents the internal investigative steps and evidentiary thresholds that precede formal reporting. While machine learning has previously been applied to insurance fraud (Aslam et al., 2022; du Preez et al., 2025), and to money laundering in banking (Jullum et al., 2020; Rocha-Salazar et al., 2021; Johannesen & Jullum, 2025), the present study develops machine learning for money laundering detection in insurance. Importantly, we broaden the auxiliary training signal beyond laundering itself, leveraging examples of insurance fraud to improve performance. To our knowledge, this is the first demonstration of such an approach on production data from an insurance company.

In conclusion, this study shows that laundering detection and prevention in the insurance sector can be meaningfully improved through machine learning. While performance depends on how the learning problem is formulated, with models that incorporate related fraud categories yielding the strongest results, the broader finding is that machine learning models can surface most laundering cases despite extreme class imbalance. This enables insurers not only to meet reporting obligations, but also to intervene earlier in the claims process and prevent high risk payout.

Ahmad, A., & Quegan, S. (2013). Comparative Analysis of Supervised and Unsupervised Classification on Multispectral Data. *Applied Mathematical Sciences*, *7*(74), 3681–3694.

Aslam, F., Awan, T. M., & Fatima, T. (2022). *Insurance fraud detection: Evidence from artificial intelligence and machine learning*.

Bagaskara, F., & Yanti, H. (2025). Examining money laundering in the insurance industry through fraud heptagon theory perspective. *Indonesian Interdisciplinary Journal of Sharia Economics*, *8*(2).

Chen, T., & Guestrin, C. (2016). *XGBoost: A Scalable Tree Boosting System*.

du Preez, A., Bhattacharya, S., Beling, P., & Bowen, E. (2025). Fraud detection in healthcare claims using machine learning: A systematic review. *Artificial Intelligence in Medicine*.

EU Directive 2024/1640, Art. 24 (2024). https://eur-lex.europa.eu/eli/dir/2024/1640/oj/eng

FATF. (2014). *Guidance for a Risk Based Approach: The Banking Sector*. Financial Action Task Force.

FATF. (2025, oktober 1). *The FATF Recommendations*. https://www.fatf-gafi.org/en/publications/Fatfrecommendations/Fatf-recommendations.html

Geirhos, R., Jacobsen, J.-H., Michaelis, C., Zemel, R. S., Brendel, W., Bethge, M., & Wichmann, F. A. (2020). Shortcut learning in deep neural networks. *Nature Portfolio (Springer Nature)*, *1*(11), 665–673.

Gottschalk, P. (2024). Money laundering prevention: The challenge of insurance termination for outlaw biker gangs' club houses. *Journal of Money Laundering Control*, *27*(6), 1040–1060.

Grinsztajn, L., Oyallon, E., & Varoquaux, G. (2022). *Why do tree-based models still outperform deep learning on tabular data? 35*, 507–520. https://doi.org/10.52202/068431-0037

Hayble-Gomes, E. (2023). The use of predictive modeling to identify relevant features for suspicious activity reporting. *Journal of Money Laundering Control*, *36*(4), 806–830.

*Insurance Authority inspection results in enhancements to anti-money laundering controls and processes and fine*. (2024, august 2). https://www.ia.org.hk/en/infocenter/press_releases/20240802.html

Johannesen, F., & Jullum, M. (2025). Finding money launderers using heterogeneous graph neural networks. *The Journal of Finance and Data Science*, *11*, 100175.

Jullum, M., Løland, A., Huseby, R. B., Ånonsen, G., & Lorentzen, J. P. (2020). Detecting money laundering transactions with machine learning. *Journal of Money Laundering Control*, *23*(1), 173–186.

Kennedy, R. K. L., Salekshahrezaee, Z., Villanustre, F., & Khoshgoftaar, T. M. (2023). Iterative cleaning and learning of big highly imbalanced fraud data using unsupervised learning. *Journal of Big Data*, *10*.

Khoei, T. T., & Kaabouch, N. (2023). A Comparative Analysis of Supervised and Unsupervised Models for Detecting Attacks on the Intrusion Detection Systems. *Information*, *14*.

KPMG. (2014). *Global Anti-Money Laundering Survey*. KPMG International Cooperative.

Law Commission. (2019). *Anti-money laundering: The SARs regime*. UK Government.

Levi, M. (2006). Money Laundering. *Crime and Justice*, *34*(1), 289–375.

Levi, M. (2019). Anti-Money Laundering: An Inquiry into a Disciplinary Transnational Legal Order. *UC Irvine Journal of International, Transnational, and Comparative Law*, *4*(1), 1–235.

Manning, C. D., Raghavan, P., & Schütze, H. (2008). *Introduction to Information Retrieval*. Cambridge University Press.

Mousavian, S., & Miah, S. J. (2025). Review of artificial intelligence-based applications for money laundering. *Intelligent Systems with Applications*.

Norwegian Director of Public Prosecutions. (2018). *Retningslinjer om heleri-, hvitvaskings- og selvvaskingssaker*.

Norwegian Ministry of Finance. (2018). *Money Laundering Act (Act of 1 June 2018 No. 23)*. https://lovdata.no/dokument/LTI/lov/2018-06-01-23

Norwegian Penal Code § 337, 2005, Nr. 337, Straffeloven.

Norwegian Penal Code § 375, 2005, Nr. 335.

Norwegian Tax Authorities. (2023, oktober 11). *Tilleggsskatt og skjerpet tilleggsskatt – forståelsen av det enkelte forhold*. https://www.skatteetaten.no/rettskilder/type/vedtak/skatteklagenemnda/tilleggsskatt-og-skjerpet-tilleggsskatt.-forstaelsen-av-det-enkelte-forhold-i-henhold-til-arbeidsgiveravgift-og-nedre-belopsgrense-for-fastsettelse-av-tilleggsskatt/

Ogbeide, H., Thomson, M., Gönül, M., Pollock, A., Bhowmick, S., & Bello, A. (2023). The anti-money laundering risk assessment: A probabilistic approach. *Journal of Business Research*.

Oztas, B., Cetinkaya, D., Adedoyin, F., Budka, M., & Gokhan, A. (2024). Transaction monitoring in anti-money laundering: A qualitative analysis and points of view from industry. *Future Generation Computer Systems*.

Pedregosa, F., Varoquaux, G., & Gramfort, A. (2011). Scikit-learn: Machine Learning in Python. *Journal of Machine Learning Research*, *12*, 2825–2830.

Proceeds of Crime Act 2002 (2002). https://www.legislation.gov.uk/ukpga/2002/29/contents

Reite, J. E., Karlsen, J., & Westgaard, E. G. (2025). Improving client risk classification with machine learning to increase anti-money laundering detection efficiency. *Journal of Money Laundering Control*, *28*(1), 93–107.

Rocha-Salazar, J.-J., Segovia-Vargas, M.-J., & Camacho-Miñano, M.-M. (2021). Money laundering and terrorism financing detection using neural networks and an abnormality indicator. *Expert Systems With Applications*, *169*.

Rosenblatt, M., Tejavibulya, L., Jiang, R., Noble, S., & Scheinost, D. (2024). Data leakage inflates prediction performance in connectome-based machine learning models. *Nature Communications*, *15*.

Sur, M., Hall, J. C., Brandt, J., Astell, M., Poessel, S. A., & Katzner, T. E. (2023). Supervised versus unsupervised approaches to classification of accelerometry data. *Ecology and Evolution*, *13*.

Thanasegaran, H., & Shanmugam, B. (2008). Exploitation of the insurance industry for money laundering: The Malaysian perspective. *Journal of Money Laundering Control*, *11*(2), 135–145.

U.S. Department of the Treasury. (2024). *2024 National Money Laundering Risk Assessment*. U.S. Government.

U.S. Government Accountability Office. (2024). *Anti Money Laundering: Better Information Needed on Suspicious Activity Reports' Use*. United States Government Accountability Office. ce.

## Appendix

*Appendix A: Three anonymised investigations*

This appendix presents three anonymised laundering investigations illustrating how suspicion of money laundering may arise in connection with insurance claims. In each case, the process began with a registered insurance claim, developed into an insurance fraud investigation, and ultimately led to a laundering investigation. The investigation summaries are based on internal documentation and describe the initial trigger for laundering suspicion, the investigative steps undertaken, the alternative explanations considered, and the final decision to submit a suspicious activity report. The examples aim to provide insight into how laundering concerns materialise in practice and how the reporting threshold is applied.

For interpretative context, guidelines from the Norwegian Director of Public Prosecutions indicate that the distinction between ordinary and aggravated money laundering rests on an overall assessment, where a natural starting point is whether the proceeds are of significant value. Proceeds above roughly USD 10,000 are considered significant (Norwegian Director of Public Prosecutions, 2018). Based on the amounts involved, the conduct described in investigations 1 and 2 would likely qualify as aggravated money laundering if proven in court, whereas investigation 3 would likely qualify as ordinary money laundering. Comparable practice from the Norwegian tax authorities suggests that investigation 3 would be handled through administrative penalties rather than criminal prosecution (Norwegian Tax Authorities, 2023).

*Investigation 1: Suspicion of organised vehicle crime*

A claim was filed for a damaged sports car, approximately valued at USD 80,000. During the routine loss assessment, vehicle part prices were obtained to determine whether repair would exceed the insured value. In this process, inconsistencies were detected between the vehicle's interior and equipment and the chassis number recorded in official registers. Technical inspection revealed signs of manipulation of the engraved chassis number in the engine compartment, including repainting and irregular typography. Electronic control systems contained a different chassis reference that better corresponded to the installed interior configuration.

These findings led to the initiation of a fraud and laundering investigation under the hypothesis that the vehicle may have been assembled using stolen parts and that the chassis number had been manipulated to conceal its origin. Further examination of the vehicle's administrative history revealed multiple ownership changes over a short period among a limited network of individuals and companies in Norway, with the vehicle originally imported from Asia. Technical verification from The Norwegian Public Roads Administration clarified that the electronic chassis reference corresponded to a vehicle internationally reported as stolen.

The combined indicators supported the hypothesis that the vehicle formed part of organised vehicle crime. A plausible explanation is that a stolen vehicle identity was attached to another chassis, or that components from different vehicles were combined to conceal origin. The repeated transfers within a closed network may have been used to obscure the vehicle's origin before it was resold or claimed under insurance.

Alternative explanations such as registration error or legitimate part replacement are difficult to reconcile with deliberate physical manipulation and the confirmed international alert linked to the electronic identifier. The threshold for serious suspicion was considered met, and a suspicious activity report was submitted to authorities.

*Investigation 2: Insurance fraud with indicators of illicit origin*

A customer filed a claim for allegedly stolen items valued at approximately USD 90,000, reporting that a burglary had taken place. During inspection, the damage to the door appeared inconsistent with forced entry, and the case was referred for fraud investigation. The investigation revealed that identical items had previously been reported stolen, several of the claimed goods had been advertised for sale online, and an external locksmith concluded that the door damage had been fabricated. Furthermore, most of the claimed items lacked purchase documentation. When questioned about this, the customer stated that many of the items had been received as gifts. However, the customer did not consent to Fremtind contacting friends and relatives to testify that they had given the gifts. On this basis, Fremtind rejected the claim and filed a police report for insurance fraud.

The absence of documentation for goods of such value also triggered a laundering investigation. The working hypothesis was that the assets may have been acquired through undeclared income or other criminal sources. A simplified private consumption analysis was conducted using available tax and debt registry data. The analysis indicated negative household liquidity, with reported income and borrowing patterns inconsistent with the acquisition of goods of this value.

The financial indicators were inconsistent with the declared assets, and the apparent fabricated burglary further weakened the credibility of the customer's explanation. Alternative explanations, such as inheritance, gifts, or legitimate resale prior to the reported burglary, were possible but not supported by documentation or testimonies. The private consumption analysis could have been refined by requesting bank statements covering the period when the goods were reportedly acquired or received, but this was not considered necessary by the responsible investigator. As the suspicion that the goods could have originated from undeclared earnings was not dispelled, Fremtind submitted a suspicious activity report to the authorities.

*Investigation 3: Blurred distinction between private and corporate assets*

A private customer submitted a claim under his home insurance policy for a damaged television valued at approximately USD 2,000. In support of the claim, he provided an order confirmation rather than a receipt. As the order confirmation appeared cropped and manipulated, the case was referred for fraud investigation. The retailer was subsequently contacted to verify the order and confirmed that the purchase was genuine. However, it had been registered not to the customer personally but to a company he was associated with.

This gave rise to the hypothesis that the asset may not have been properly transferred from the company to the customer for private use, potentially in breach of tax or accounting rules. Fremtind requested bank documentation of such a transfer. The customer did not provide bank documentation but instead produced an internal company invoice marked as paid, insisting that the transaction had been an internal financial settlement without clarifying how this was recorded in the accounts.

The television may legitimately have been taken into private use. However, after being given a second opportunity to provide adequate documentation following the initial inconsistencies, the customer did not present evidence showing that the transfer had been properly recorded in the company's accounts or that any VAT adjustments had been made. Although the suspicion of fabricated documentation was dispelled, the uncertainty concerning ownership and transfer could not be resolved. Fremtind therefore submitted a suspicious activity report to the authorities.

*Appendix B: Methodological details*

This appendix provides supplementary details on the experimental setup. It reports the class distributions for each outer and inner split in the nested time-series cross-validation scheme, as well as the hyperparameter grids used in the experiments. While the proprietary nature of the data prevents full reproducibility, these details aim to improve transparency and document how the models were trained and evaluated.

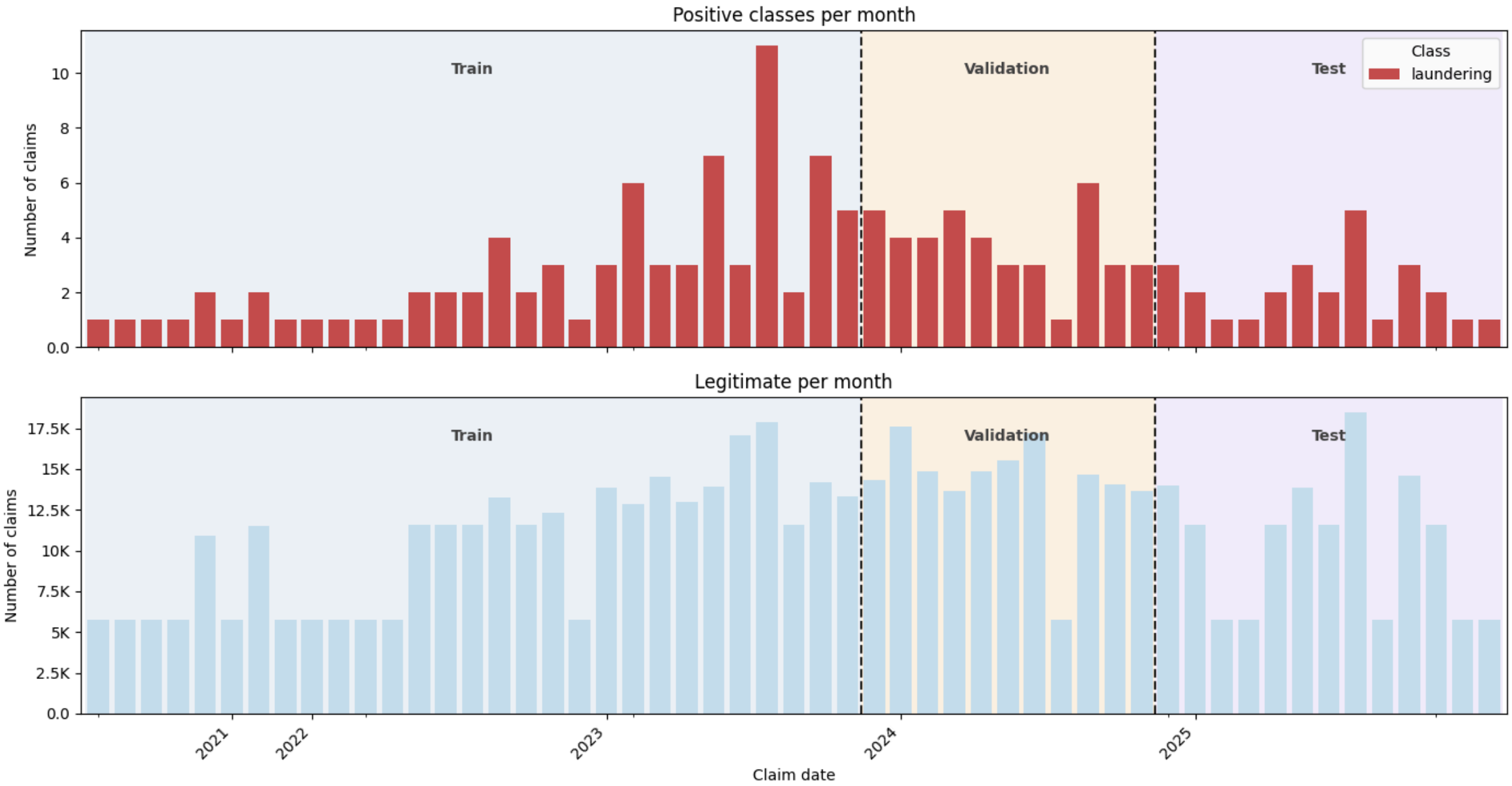

**Figure B1:** Schematic of the time-series nested CV pipeline for learning setup Binary-Laundering, using dataset realisation 1, outer fold 3, and inner fold 1. Each claim in the figure is associated with a unique customer and the distribution of dates are roughly matched between the laundering class and the legitimate class.

| ***Outer fold*** | ***Split*** | ***Legitimate*** | ***Fraud investigation*** | ***Fraud*** | ***Laundering*** |
|---|---|---|---|---|---|
| *0* | *train+val* | *160 497* | *0* | *0* | *32* |
| *0* | *test* | *93 872* | *0* | *0* | *29* |
| *1* | *train+val* | *254 369* | *0* | *0* | *61* |
| *1* | *test* | *80 604* | *0* | *0* | *29* |
| *2* | *train+val* | *334 973* | *0* | *0* | *90* |
| *2* | *test* | *111 894* | *0* | *0* | *29* |
| *3* | *train+val* | *446 867* | *0* | *0* | *119* |
| *3* | *test* | *140 902* | *0* | *0* | *29* |

***Table B1:*** *Class counts for outer fold 3 for learning setup Binary-Laundering and dataset realisation 1.*

| ***Outer fold*** | ***Inner fold*** | ***Split*** | ***Legitimate*** | ***Fraud investigation*** | ***Fraud*** | ***Laundering*** |
|---|---|---|---|---|---|---|
| *3* | *0* | *train* | *189 490* | *0* | *0* | *41* |
| *3* | *0* | *validation* | *108 305* | *0* | *0* | *39* |
| *3* | *1* | *train* | *297 795* | *0* | *0* | *80* |
| *3* | *1* | *validation* | *149 072* | *0* | *0* | *39* |

***Table B2:*** *Class counts for inner folds within outer fold 3 for learning setup Binary-Laundering and dataset realisation 1. The inner splits follow an expanding time window and are constructed chronologically based on claim date.*

| ***Outer fold*** | ***Split*** | ***Legitimate*** | ***Fraud investigation*** | ***Fraud*** | ***Laundering*** |
|---|---|---|---|---|---|
| *0* | *train+val* | *198 814* | *181* | *648* | *32* |
| *0* | *test* | *945 49* | *351* | *233* | *29* |
| *1* | *train+val* | *293 363* | *532* | *881* | *61* |
| *1* | *test* | *72 515* | *263* | *137* | *29* |
| *2* | *train+val* | *365 878* | *795* | *1 018* | *90* |
| *2* | *test* | *119 706* | *425* | *225* | *29* |
| *3* | *train+val* | *485 584* | *1 220* | *1 243* | *119* |
| *3* | *test* | *171 061* | *541* | *323* | *29* |

***Table B3:*** *Class counts for outer fold 3 for learning setups Binary-Investigation and Multiclass and dataset realisation 1.*

| ***Outer fold*** | ***Inner fold*** | ***Split*** | ***Legitimate*** | ***Fraud investigation*** | ***Fraud*** | ***Laundering*** |
|---|---|---|---|---|---|---|
| *3* | *0* | *Train* | *223 716* | *272* | *705* | *41* |
| *3* | *0* | *Val* | *109 596* | *403* | *251* | *39* |
| *3* | *1* | *Train* | *333 312* | *675* | *956* | *80* |
| *3* | *1* | *Val* | *152 272* | *545* | *287* | *39* |

***Table B4:*** *Class counts for inner folds within outer fold 3 for learning setups Binary-Investigation and Multiclass and dataset realisation 1.*

We below document the hyperparameter grids used across all experiments. For comparability across learning formulations, the total number of evaluated model configurations was kept constant. Each learning setup evaluates 80 configurations, computed as 4 XGBoost hyperparameter combinations multiplied by the relevant class-weight grid and probability combination options. Early stopping with 50 rounds was applied in all experiments, with a maximum of 5,000 boosting rounds.

*Table B5: XGBoost hyperparameter grid used in all learning*

| *Setting* | *Grid* |
|---|---|
| *Learning rate* | *{0.005, 0.02}* |
| *Max depth* | *{3, 5}* |

***Table B5***: *XGBoost hyperparameter grid used in all learning setups. Parameters not listed were set to their default values. The grid yields 4 model configurations.*

| ***Binary model*** | | |
|---|---|---|
| ***Positive weight*** | ***Positive weight (continued)*** | ***Positive weight (continued)*** |
| *5* | *135* | *2200* |
| *10* | *200* | *3200* |
| *15* | *300* | *4500* |
| *25* | *450* | *5200* |
| *40* | *675* | *5800* |
| *60* | *1000* | *6400* |
| *90* | *1500* | |

***Table B6***: *Sample weight grid used in binary and multiclass learning setups.*

| ***Multiclass model*** | | |
|---|---|---|
| ***Fraud_investigation_weight*** | ***Fraud_weight*** | ***Laundering_weight*** |
| *1* | *1* | *1* |
| *1* | *1* | *10* |
| *1* | *1* | *15* |
| *2* | *2* | *25* |
| *3* | *3* | *40* |
| *5* | *5* | *60* |
| *7* | *7* | *90* |
| *10* | *10* | *135* |
| *15* | *15* | *200* |
| *20* | *20* | *300* |
| *30* | *30* | *450* |
| *45* | *45* | *675* |
| *70* | *70* | *1000* |
| *100* | *100* | *1500* |
| *150* | *150* | *2200* |
| *200* | *200* | *3200* |
| *250* | *250* | *4500* |
| *280* | *280* | *5200* |
| *300* | *300* | *5800* |
| *320* | *320* | *6400* |

***Table B7**: Sample-weight grids used in the binary and multiclass learning setups. The binary grid contains 20 weight configurations. The multiclass grid contains 20 predefined (risk_weight, laundering_weight) pairs.*

| ***Setup*** | ***XGBoost grid*** | ***Weight grid*** | ***Total*** |
|---|---|---|---|

| *Binary-Laundering* | *4* | *20* | *80* |
|---|---|---|---|
| *Binary-Investigation* | *4* | *20* | *80* |
| *Multiclass* | *4* | *20* | *80* |

***Table B8***: *Total configurations per learning setup*

*Appendix C: Additional results*

This appendix reports detailed performance results across realisations and outer folds for the Budget-Weighted Capture Rate (BWCR) and Average Precision. All reported values are percentages. Each experiment includes 16 evaluations (4 realisations × 4 outer folds).

| ***Experiment*** | ***Realisation 1*** | ***Realisation 2*** | ***Realisation 3*** | ***Realisation 4*** | ***Across realisations*** |
|---|---|---|---|---|---|
| *Binary–Laundering* | *54.74* | *50.65* | *53.45* | *53.88* | *53.18* |
| *Binary–Investigation* | *64.87* | *64.44* | *65.09* | *64.66* | *64.76* |
| *Multiclass* | *57.33* | *59.05* | *58.84* | *56.03* | *57.81* |

***Table C1***: *BWCR — Per Realisation Averages (%)*

| ***Experiment*** | ***Realisation 1*** | ***Realisation 2*** | ***Realisation 3*** | ***Realisation 4*** | ***Across seeds*** |
|---|---|---|---|---|---|
| *Binary–Laundering* | *1.01* | *1.44* | *1.50* | *1.75* | *1.42* |
| *Binary–Investigation* | *2.29* | *1.95* | *2.84* | *2.28* | *2.34* |
| *Multiclass* | *2.09* | *1.96* | *2.45* | *2.20* | *2.18* |

***Table C2***: *Average Precision — Per Realisation Averages (%)*

| ***Experiment*** | ***Fold 0*** | ***Fold 1*** | ***Fold 2*** | ***Fold 3*** |
|---|---|---|---|---|
| *Binary–Laundering* | *0.54±0.11* | *0.65±0.14* | *1.49±0.88* | *3.01±1.08* |
| *Binary–Investigation* | *4.00±1.28* | *2.47±0.25* | *0.99±0.11* | *1.90±0.26* |
| *Multiclass* | *0.45±0.08* | *1.27±0.29* | *3.50±1.19* | *3.48±0.96* |

***Table C3***: *Average Precision — Per Outer Fold Averages (%)*

*Appendix D: Lovdata Search*

A search was conducted in Lovdata, Norway's public database of court decisions, in February 2026 using the keywords "fremtind*" and "hvitvask*". The search returned ten results, of which five were court rulings. None of the identified rulings involved the prosecution or conviction of a Fremtind customer for money laundering.

| ***Reference*** | ***Court/Body*** | ***Brief description*** |
|---|---|---|
| *LE-2023-68349* | *Eidsivating lagmannsrett* | *Theft; reimbursement of deductible to Fremtind* |
| *LB-2021-84438* | *Borgarting lagmannsrett* | *Theft; compensation to Fremtind* |
| *LB-2024-184049* | *Borgarting lagmannsrett* | *Insurance fraud against Fremtind* |
| *LE-2023-103081* | *Eidsivating lagmannsrett* | *Theft; compensation to Fremtind* |
| *TVIN-2023-164632* | *Vestre Innlandet tingrett* | *Theft; compensation to Fremtind* |
| *TOBYF-2020-117337* | *Oslo byfogdembete* | *Disclosure between financial institutions* |
| *FinKN-2025-714* | *Finansklagenemnda* | *Insurance payout dispute (Rolex watches and handbag)* |
| *FinKN-2025-147* | *Finansklagenemnda* | *Marketing Act issue (broker liability insurance)* |
| *FinKN-2025-1109* | *Finansklagenemnda* | *Legal aid coverage dispute* |
| *(Advokatloven amendment)* | *Legislative document* | *Amendment to the Lawyers Act* |